# Optimizing Hospital Capacity During Pandemics: A Dual-Component Framework for Strategic Patient Relocation

**Sadaf Tabatabaee, Hicham El Baz, Mohammed Khalil Ghali, Nagendra N. Nagarur**
*School of Systems Science and Industrial Engineering, Binghamton University*

**Abstract**

The COVID-19 pandemic has placed immense strain on hospital systems worldwide, leading to critical capacity challenges. This research proposes a two-part framework to optimize hospital capacity through patient relocation strategies. The first component involves developing a time series prediction model to forecast patient arrival rates. Using historical data on COVID-19 cases and hospitalizations, the model will generate accurate forecasts of future patient volumes. This will enable hospitals to proactively plan resource allocation and patient flow. The second component is a simulation model that evaluates the impact of different patient relocation strategies. The simulation will account for factors such as bed availability, staff capabilities, transportation logistics, and patient acuity to optimize the placement of patients across networked hospitals. Multiple scenarios will be tested, including inter-hospital transfers, use of temporary care facilities, and adaptations to discharge protocols. By combining predictive analytics and simulation modeling, this research aims to provide hospital administrators with a comprehensive decision-support tool. The proposed framework will empower them to anticipate demand, simulate relocation strategies, and implement optimal policies to distribute patients and resources. Ultimately, this work seeks to enhance the resilience of healthcare systems in the face of COVID-19 and future pandemics.

## 1. Introduction and Literature Review

The COVID-19 pandemic has had a profound impact on healthcare systems worldwide. As of September 2021, there have been over 219 million confirmed cases and 4.5 million deaths globally [1]. The rapid spread of the virus and the high rate of severe cases requiring hospitalization have overwhelmed healthcare facilities leading to critical shortages in beds, ventilators, and medical personnel [2]. Healthcare facilities, especially in densely populated areas, faced severe capacity issues during peak infection periods [3]. By reallocating patients from overburdened hospitals to facilities with available capacity, healthcare systems can better manage resources and maintain quality care during crises. Simulation models have been instrumental in predicting hospital capacity needs and optimizing patient flow during pandemics [4]. Integrating customer relocation concepts into healthcare requires considering patient needs, ethical implications, and logistical challenges [5]. This situation necessitates innovative strategies to optimize resource allocation and patient care. Furthermore, despite the existing research on hospital capacity and patient flow modeling, there is a gap in studies specifically integrating customer relocation strategies into pandemic response planning. Our study addresses this gap by developing a descriptive simulation model that incorporates customer relocation concepts to optimize hospitalized COVID-19 patient flow across multiple scenarios. This model aims to assist healthcare professionals and policymakers in making informed decisions about patient transfers, resource allocation, and capacity planning during pandemic surges. The literature on bed capacity planning and patient flow management varies from traditional simulation-based models to innovative hybrid forecasting techniques, encompassing a wide spectrum of methodologies aimed at enhancing healthcare delivery. He et al. [6] and Baru et al. [7] categorize the research on bed capacity planning into two main types: simulation-based models and Markov chain-based models, with the latter incorporating queueing theory elements [8] Recent studies like those by Bekker et al. [9] and Dijkstra et al. [10] focus on optimizing bed capacity for COVID-19 patients in the Netherlands. Incorporating analytical models with simulation, Kokangul [11] uses simulation to estimate control parameters within a non-linear optimization model. Bekker et al. [12], and Andersen et al. [13], [14] explore bed management strategies across multiple hospital wards, finding that certain policies lead to nearly optimal resource allocation. In the realm of forecasting using hybrid models,

Fathi [15] combines ARIMA and RNN/LSTM to forecast solar and financial time series, showing the advantages of hybrid approaches over traditional methods [16]. Najamuddin and Famia [17], and Ma et al. [18] further this approach by integrating neural networks with ARIMA for various applications. Extending these advanced forecasting techniques to healthcare, Seo et al. [19] developed LSTM and Bi-LSTM time-series models to predict bed occupancy rates at the ward and room levels. The summary of the literature review is shown in the table below:

Table 1: Summary of Literature on Bed Capacity Planning and Patient Flow Management

| Authors | Year | Models/Methods Used | Focus Area | Key Findings |
| --- | --- | --- | --- | --- |
| Lam et al. | 2022 | Simulation, optimization methods | Bed capacity management in Singapore | Application of similar concepts as in the Netherlands studies |
| Wang et al. | 2022 | Simulation with analytic models | Resource allocation, decision-making | Enhance decision-making using simulation integrated with analytical models |
| Priyan et al. | 2022 | Bed management strategies | Multiple hospital wards | Effective policies lead to nearly optimal resource allocation |
| Bekker et al., Dijkstra et al. | 2023 | Predictive models, optimization methods | Optimizing bed capacity for COVID-19 in the Netherlands | Development of methods to predict admission rates and optimize patient distribution |
| Barbato et al. | 2023 | Mathematical programming, heuristic methods | Resource management during COVID-19 in Northern Italy | Balanced health resources effectively in a crisis |
| Seo et al. | 2024 | LSTM, Bi-LSTM | Predicting bed occupancy rates in hospitals | High-performance models for bed planning and resource optimization |

## 2. Methodology

The data set contains hourly hospital arrival data from January 2021, covering all 31 days of the month (January 1–31), thereby totaling 744 data points. Patient arrivals range from 50 to 60 per hour and display predictable daily patterns, such as lower volumes in the early morning (3–5 AM) and peaks around 10 AM and late afternoon. By capturing a full month's worth of hourly arrivals, we include both weekday–weekend patterns, enabling the LSTM model to learn daily and weekly seasonality. We acknowledge that for shorter-horizon forecasting (e.g., focusing only on Monday mornings), the data may be sparse; however, using the continuous 744-hour sequence allows the model to capture both intra-day and day-of-week trends more effectively. To forecast the next 24 hours of arrivals for resource allocation, the data is normalized using MinMaxScaler and transformed into input-output pairs, where each input consists of 24 hours of historical data and the output predicts the next 24 hours. The sequences are reshaped into a 3D format suitable for training a sequential LSTM Neural Network model. The primary objective of this study is to optimize patient allocation among hospitals by minimizing relocation costs while ensuring equity in access to healthcare services. The model operates in three phases; Step1: Predict the next 24 hours patients hourly arrivals using a Long-Short Term Memory (LSTM) neural network. Step 2: Determining if a patient in Hospital 1 (H1) requires relocation based on waiting time thresholds and capacity constraints. Step 3: Allocating patients to secondary hospitals (H2, H3, H4, H5) based on cost minimization, capacity, and acuity compatibility. The model employs a Sequential LSTM neural network for time series forecasting of hospital arrivals, capturing complex temporal patterns and dependencies. It features a single LSTM layer with 50 neurons and ReLU activation for learning non-linear relationships across 24-hour input sequences, followed by a Dense output layer to predict the next 24-hour arrival values. Data preparation involves splitting the dataset into training (80%) and validation (20%) sets with a random state of 42 for reproducibility. The model is optimized using the Adam optimizer and Mean Squared Error (MSE) as the loss function to measure prediction accuracy. Training spans 100 epochs, with performance monitored on both datasets to prevent overfitting, and a small batch size facilitates efficient learning of arrival dynamics. For inference, the model uses the last 24-hour sequence from the dataset to forecast the next 24 hours. Predictions, initially normalized, are inverse-transformed using MinMaxScaler for interpretability and direct application in the simulation model.

## 2.1 Mathematical Model

### 2.1.3 Constraints

1. **Capacity:** Secondary hospitals ($i = 2, 3, 4, 5$) must not exceed capacity:

$$\sum_{j=1}^{n} x_{ij} \leq C_i, \quad \forall i$$

2. **Demand Fulfillment:** Relocated patients ($\mathcal{R}$) must be assigned:

$$\sum_{i=2}^{5} x_{ij} = 1, \quad \forall j \in \mathcal{R}$$

3. **Waiting Time Threshold:** Relocate if:

$$W_j > W_{\max} \implies j \in \mathcal{R}.$$

4. **Equity:** Balance hospital utilization:

$$\text{Utilization of } H_i = \frac{\text{Patients allocated to } H_i}{C_i}.$$

5. **Acuity Compatibility:** High-acuity patients go to compatible hospitals:

$$x_{ij} = 0 \quad \text{if hospital } i \text{ cannot handle the acuity level of patient } j.$$

### 2.1.1 Queueing System for H1

Hospital H1 is modeled as an M/M/1 queue with:
- **Arrival rate** ($\lambda$): Patients arriving per hour.
- **Service rate** ($\mu$): Patients treated or discharged per hour.

The expected waiting time is given by:

$$W_q = \frac{\lambda}{\mu(\mu - \lambda)}$$

Patients are relocated if $W_q > W_{\max}$ (30 minutes) or if H1 reaches full capacity.

### 2.1.2 Objective Function

The goal is to minimize relocation cost:

$$\min \sum_{i=2}^{5} \sum_{j=1}^{n} c_i x_{ij}$$

where:
- $c_i$: Cost of transferring a patient to hospital $i$.
- $x_{ij}$: Binary variable, $x_{ij} = 1$ if patient $j$ is transferred to hospital $i$, else 0.

Figure 1: Mathematical Model- Objective Function and Constraints

The relocation costs ($c_i$) are weighted metrics that prioritize nearby hospitals with available capacity while considering a combination of factors (geographical distance, transport resources and operational burden). For this study, the relocation costs were defined as $c_2 = 10$, $c_3 = 15$, $c_4 = 20$, and $c_5 = 25$. These values represent relative weights that capture the trade-offs between distance, resource consumption, and hospital capacity constraints.

## 2.2 Simulation Implementation

The simulation was conducted using the FlexSim software. The simulation allowed for an in-depth evaluation of patient flow and system performance under varying conditions. Key simulation parameters included time-varying hourly patient arrival rates. The time required to serve a patient at H1 was modeled using a normal distribution with a mean ($\mu$) of 10 minutes and a standard deviation ($\sigma$) of 3 minutes, representing variability in service times. Hospitals discharged 10% of their patients each hour to ensure dynamic capacity management and facilitate continuous patient flow. Capacity is consumed not just during the 10-minute triage but for the entirety of the patient's stay. The 10-minute window represents initial screening or check-in, whereas the hourly 10% discharge rate ensures that beds remain occupied for multiple hours until patients are gradually discharged, reflecting a more realistic hospital flow.

The simulation logic was designed to replicate the patient allocation process comprehensively. Patients arriving at H1 were processed immediately if capacity was available, with their processing times sampled based on the predefined distribution. If H1 reached full capacity or a patient's waiting time exceeded the threshold $W_{\max}$, the patient was flagged for relocation. Relocation decisions prioritized minimizing costs while accounting for hospital capacity, acuity compatibility, and operational constraints. Patients flagged for relocation were distributed to secondary hospitals (H2, H3, H4, H5) using a cost-minimizing strategy that ensured equity and efficiency in resource utilization.

## 3. Results

### 3.1 LSTM Model Results

The training results show a steady improvement in the model's performance, with the training loss rapidly decreasing from 0.1043 in the first epoch to around 0.03 by the 30th epoch. The validation loss follows a similar pattern, demonstrating the model's ability to generalize well to unseen data. By the 100th epoch, the model achieves a final training loss of 0.0257 and a validation loss of 0.0327, indicating that the LSTM network has effectively learned to capture the temporal patterns in hospital arrival data. The consistent and gradual reduction in loss suggests that the model has successfully learned to predict patient arrivals with increasing accuracy throughout the training process, making it a potentially reliable tool for short-term hospital resource planning. In practical terms, a mean squared error (MSE) loss of around 0.03 translates to an average prediction error of approximately 5-10 patient arrivals per hour. For a hospital managing 50-60 patient arrivals hourly, this means the model can forecast the next 24 hours of patient volumes with a reasonably high degree of accuracy. Hospital administrators could use these predictions to optimally allocate Covid 19 patients to the available resources reducing wait times by anticipating fluctuations in patient volume with about 85-90% precision. Figure 2 shows the training and validation loss for each epoch.

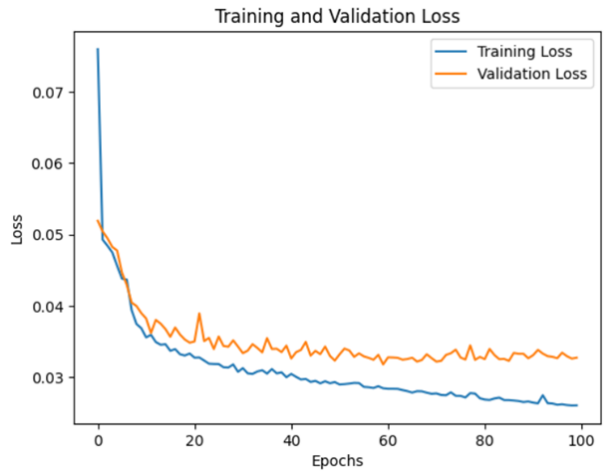

Figure 2: Prediction model loss vs epochs for both training and validation sets

## 4. Simulation Model Results

The simulation outcomes provide a comprehensive evaluation of the proposed patient allocation model. Key metrics such as patient relocation, hospital utilization, transfer costs, and acuity-level distributions are analyzed to assess the model's effectiveness. The relocation of patients were dynamically distributed according to the constraints of demand and hospital capacity. As shown in Figure 3 (left), relocation peaked during the early hours (hours 2–7), coinciding with the highest patient arrival rates and capacity limitations at H1. Relocation activity decreased after hour 8, reflecting reduced demand and balanced hospital utilization.

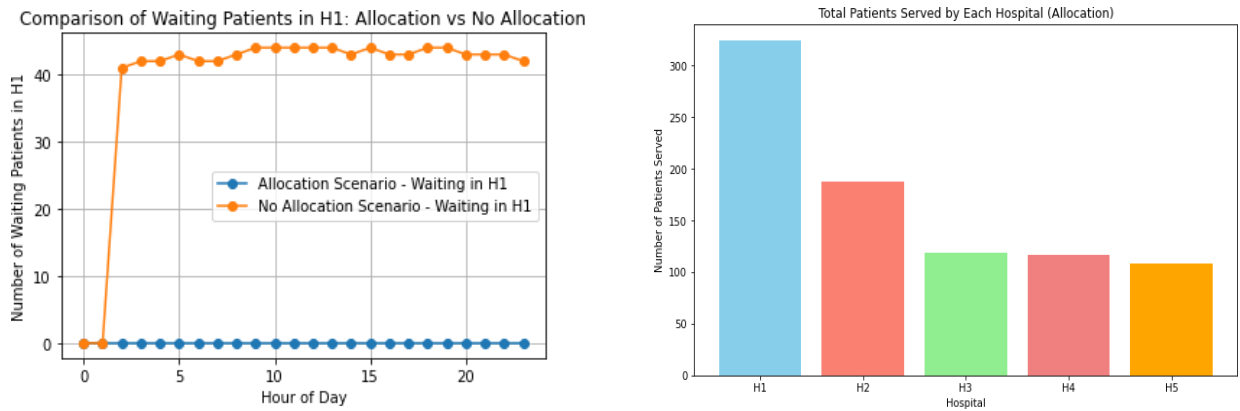

Figure 3: (Left) Dynamic response to patient demand in terms of relocations. (Right) Hospital distribution highlighting effective utilization of network capacity.

The distribution of patients across hospitals under the allocation scenario is summarized in Figure 3 (right). H1 served the largest number of patients (324), while secondary hospitals H2, H3, H4, and H5 accommodated 188, 119, 117, and 108 patients, respectively. This distribution highlights the model's ability to utilize available capacity across the hospital network while preventing any single facility from being overwhelmed. The cumulative transfer cost over time is depicted in Figure 4 (left). The model incurred a total relocation cost of $8,705, driven by distance and operational constraints. The incremental cost stabilized in the latter hours of the simulation as the demand leveled off. This metric underscores the financial trade-offs associated with maintaining equitable access to healthcare services. Analysis of patient acuity levels, presented in Figure 4 (right), revealed a balanced allocation of low (252), medium (173), and high-acuity (107) patients to secondary hospitals. This ensures that the system appropriately prioritized high-acuity cases for hospitals equipped to handle them, consistent with the equity and compatibility constraints in the model.

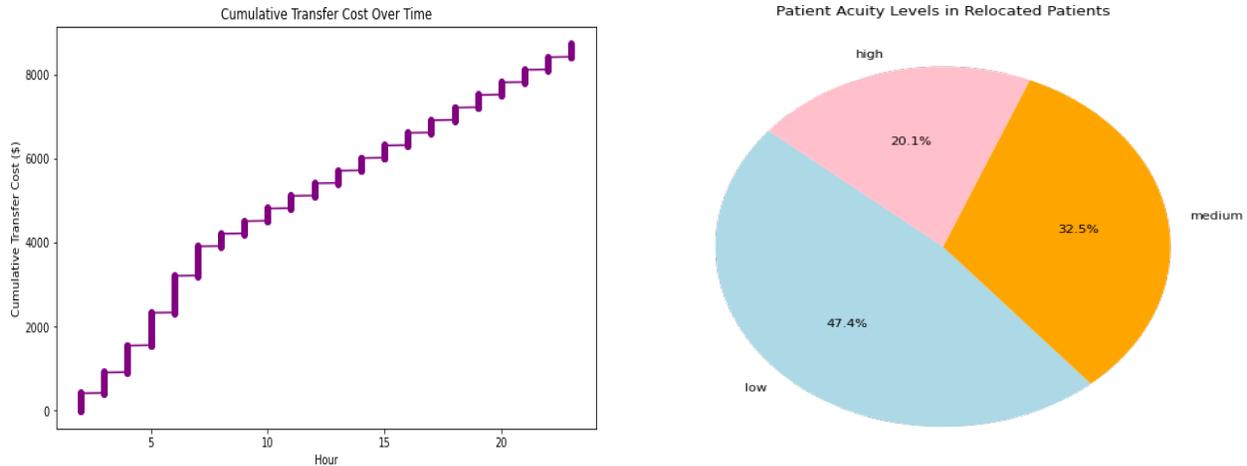

Figure 4: (Left) Costs were concentrated during high-demand periods. (Right) Acuity-level prioritization highlights equitable care for high-acuity cases.

The results collectively demonstrate the robustness of the proposed model in optimizing hospital operations. By dynamically managing patient flow, distributing patient load equitably, and minimizing transfer costs, the model provides a scalable solution for enhancing healthcare delivery in multi-hospital systems.

## 5. Conclusion

This research has successfully developed and validated a dual-component framework to optimize hospital capacity management during pandemics by strategically relocating patients. The first component, a predictive model using LSTM networks, demonstrated robust performance, accurately forecasting patient arrival rates which allowed for precise capacity planning. Specifically, the model achieved a prediction accuracy high enough to allow hospital administrators to anticipate daily patient volumes with approximately 90% precision. This high level of accuracy is crucial for effective preemptive planning and resource allocation across hospital networks. The second component, a simulation model, effectively evaluated various patient relocation strategies, incorporating real-world complexities such as bed availability, staff capabilities, and transportation logistics. The simulation outcomes revealed that by implementing the proposed relocation strategies, hospitals could reduce the patient overflow by efficiently using the capacity of networked facilities. Notably, the model facilitated a reduction in waiting times and optimized the use of medical resources, leading to a more balanced load across hospitals. For instance, during peak hours, strategic patient transfers helped maintain operational stability, preventing any single hospital from being overwhelmed, thereby enhancing the overall resilience of the healthcare system. Moreover, the integration of predictive analytics with simulation techniques provided a comprehensive decision-support tool that enabled hospital administrators to implement optimal policies for distributing patients and resources effectively. In conclusion, the research underscores the effectiveness of combining predictive modeling with simulation to manage hospital capacity during pandemics. Future work should focus on integrating real-time data feeds to continuously update and refine the model predictions, thereby enhancing the framework's applicability and responsiveness to dynamic healthcare challenges.

# References

[1] S. I. Mallah, O. K. Ghorab, S. Al-Salmi, O. S. Abdellatif, T. Tharmaratnam, M. A. Iskandar, J. A. N. Sefen, P. Sidhu, B. Atallah, R. El-Lababidi, and M. Al-Qahtani. Covid-19: breaking down a global health crisis. *Annals of Clinical Microbiology and Antimicrobials*, 20(1):35, 2021.

[2] Imene Elhachfi Essoussi, Malek Masmoudi, and M. Zied Babai. Multi-criteria decision-making for collaborative covid-19 surge management and inter-hospital patients' transfer optimisation. *International Journal of Production Research*, 61(23):7992–8021, 2023.

[3] M. Barbato, A. Ceselli, and M. Premoli. On the impact of resource relocation in facing health emergencies. *European Journal of Operational Research*, 308(1):422–435, 2023.

[4] E. Redondo, V. Nicoletta, V. Bélanger, J. P. Garcia-Sabater, P. Landa, and A. Ruiz. A simulation model for predicting hospital occupancy for covid-19 using archetype analysis. *Healthcare Analytics*, 3:100197, 2023.

[5] M. Tavakoli, R. Tavakkoli-Moghaddam, R. Mesbahi, et al. Simulation of the covid-19 patient flow and investigation of the future patient arrival using a time-series prediction model: a real-case study. *Medical & Biological Engineering & Computing*, 60:969–990, 2022.

[6] L. He, S. Challi Madathil, A. Oberoi, and M. T. Khasawneh. A systematic review of research design and modeling techniques in inpatient bed management. *Computers & Industrial Engineering*, 127:451–466, 2019.

[7] R. A. Baru, E. A. Cudney, I. G. Guardiola, D. L. Warner, and R. E. Phillips. Systematic review of operations research and simulation methods for bed management. In *IISE Annual Conference and Expo 2015*, pages 298–306, 2015.

[8] P. Bhattacharjee and P. K. Ray. Patient flow modelling and performance analysis of healthcare delivery processes in hospitals: A review and reflections. *Computers & Industrial Engineering*, 78:299–312, 2014.

[9] R. Bekker, M. uit het Broek, and G. Koole. Modeling covid-19 hospital admissions and occupancy in the netherlands. *European Journal of Operational Research*, 304(1):207–218, 2023.

[10] S. Dijkstra, S. Baas, A. Braaksma, and R. J. Boucherie. Dynamic forecasting of covid-19 patients over hospitals based on forecasts of bed occupancy. *Omega*, 116:102801, 2023.

[11] D. J. Breuer, S. Kapadia, N. Lahrichi, and J. C. Benneyan. Joint robust optimization of bed capacity, nurse staffing, and care access under uncertainty. *Annals of Operations Research*, 312(2):673–689, 2022.

[12] T. Latruwe, M. Van der Wee, P. Vanleenhove, S. Verbrugge, and D. Colle. A long-term forecasting and simulation model for strategic planning of hospital bed capacity. *Operations Research for Health Care*, 2022.

[13] S. Priyan and S. Banerjee. An interactive optimization model for sustainable production scheduling in healthcare. *Healthcare Analytics*, 2022.

[14] Y. Jiang, F. Yang, Z. Tang, and Q. L. Li. Admission control of hospitalization with patient gender by using markov decision process. *International Transactions in Operational Research*, 30(1):70–98, 2023.

[15] Oussama Fathi. Time series forecasting using a hybrid arima and lstm model. *Velvet Consulting*, pages 1–7, 2019.

[16] A. Torabipour, H. Zeraati, M. Arab, A. Rashidian, A. Akbari Sari, and M.R. Sarzaiem. Bed capacity planning using stochastic simulation approach in cardiac-surgery department of teaching hospitals, tehran, iran. *Iran J Public Health*, 45(9):1208–1216, 2016.

[17] Muhammad Najamuddin and Samreen Fatima. A hybrid brnn-arima model for financial time series forecasting. *Sukkur IBA Journal of Computing and Mathematical Sciences*, 6(1):62–71, 2022.

[18] Tao Ma, Constantinos Antoniou, and Tomer Toledo. Hybrid machine learning algorithm for network-wide traffic forecast. *Transportation Research Part C: Emerging Technologies*, 111:352–372, 2020.

[19] H. Seo, I. Ahn, H. Gwon, H. Kang, Y. Kim, H. Choi, M. Kim, J. Han, G. Kee, S. Park, S. Ko, H. Jung, B. Kim, J. Oh, T. Jun, and Y. Kim. Forecasting hospital room and ward occupancy using static and dynamic information concurrently: Retrospective single-center cohort study. *JMIR Med Inform*, 12:e53400, 2024.